\documentclass[letterpaper, 10 pt, conference]{ieeeconf}  

\IEEEoverridecommandlockouts                              

\makeatletter
\let\NAT@parse\undefined
\makeatother

\usepackage{amsmath} 
\usepackage{bbm}
\usepackage{amssymb}
\usepackage{color}
\usepackage{dsfont}

\usepackage{wrapfig}
\usepackage{graphicx} 
\usepackage{subfigure} 

\usepackage{algorithm}
\usepackage{algorithmic}
\usepackage{mathtools}

\usepackage[numbers]{natbib}

\usepackage{soul} 

\renewcommand{\cite}{\citep}

 \renewcommand{\baselinestretch}{0.98}

\newenvironment{packed_item}{
\begin{itemize}
  \setlength{\itemsep}{0pt}
  \setlength{\parskip}{0pt}
  \setlength{\parsep}{0pt}
}{\end{itemize}}

\newlength{\sectionReduceTop}
\newlength{\sectionReduceBot}
\newlength{\subsectionReduceTop}
\newlength{\subsectionReduceBot}
\newlength{\abstractReduceTop}
\newlength{\abstractReduceBot}
\newlength{\captionReduceTop}
\newlength{\captionReduceBot}
\newlength{\subsubsectionReduceTop}
\newlength{\subsubsectionReduceBot}

\newlength{\horSkip}
\newlength{\verSkip}

\newlength{\figureHeight}
\title{\LARGE \bf
Synthesizing Manipulation Sequences for Under-Specified Tasks\\ using Unrolled Markov Random Fields}

\author{Jaeyong Sung, Bart Selman and Ashutosh Saxena
\thanks{Jaeyong Sung, Bart Selman and Ashutosh Saxena are with 
        the Department of Computer Science, Cornell University, Ithaca, NY.
        Email: {\tt\small \{jysung,selman,asaxena\}@cs.cornell.edu}}%
}

\DeclareMathOperator*{\argmax}{arg\,max}

\begin{document}

\maketitle
\thispagestyle{empty}
\pagestyle{empty}

\def\mathbi#1{\textbf{\em #1}}


\begin{abstract}
Many tasks in human environments require performing a sequence of
 navigation and manipulation steps involving objects. 
 In unstructured human environments, the location and configuration of the objects
  involved often change in unpredictable ways. This requires a high-level planning
strategy that is robust and flexible in an uncertain environment. We
propose a novel dynamic planning strategy, which can be trained from a
set of example sequences. High level tasks are expressed as a
sequence of primitive actions  or controllers (with appropriate
parameters). 
Our score function, based on Markov Random Field (MRF), captures the relations between environment,
controllers, and their arguments.
By expressing the environment using sets of
attributes, the approach generalizes well to unseen scenarios. 
We train the parameters of our MRF using a maximum
margin learning method. We provide a detailed empirical validation
of our overall framework demonstrating successful plan strategies for
a variety of tasks.\footnote{A preliminary version of this work was presented 
at ICML workshop on Prediction with Sequential Models, 2013 \cite{sung_learningsequences_2013}.
}
\end{abstract}


\section{Introduction}


When interacting with a robot, users often under-specify the tasks to be performed. 
For example in Figure~\ref{fig:exampletask}, when asked to \texttt{pour} something,
the robot has to infer 
which cup to pour into and 
a complete sequence of the navigation and manipulation 
steps---moving close, grasping, placing, and so on.

This sequence not only changes with the task, but also with the perceived state 
of the environment.
As an example, consider the task of a robot fetching a magazine from a desk.  
The method to perform this task varies
depending on several properties of the environment: for example,
the robot's relative distance from the magazine, the robot's relative
orientation, the thickness of the magazine, and the presence or the absence of
other items on top of the magazine. If the magazine is very thin, the
robot may have to slide the magazine to the side of the table to pick
it up. If there is a mug sitting on top of the magazine, it would have
to be moved prior to the magazine being picked up. 
Thus, especially when the details of the manipulation task are under-specified,
the success of executing the task
depends on the ability to detect the object and
on the ability to sequence the set of \emph{primitives} (navigation and manipulation controllers)
in various ways in response to the environment.

In recent years, there have been significant developments
in building low-level controllers for robots
\cite{thrun2005probabilistic} as well as in perceptual tasks such as
object detection from sensor data \cite{koppula2011semantic,jiang-hallucinatinghumans-labeling3dscenes-cvpr2013,wulenzsaxena2014_hierarchicalrgbdlabeling}.
In this work, our goal is to, given the environment and the task,
enable robots to sequence the navigation and manipulation primitives.
Manually sequencing instructions
is not scalable because of the large
variety of tasks and situations that can arise in unstructured environments.

In this work, we take an attribute-based representation of the environment,
where each object is represented with a set of 
attributes, such as their size, shape-related information, presence of
handles, and so forth. 
For a given task, there are often multiple objects with similar
functions that can be used to accomplish the task,
and humans can naturally reason and choose the most
suitable object for the given task \cite{kemp2010learning}. 
Our model, based on attribute representation of objects,
is similarly capable of choosing the most
suitable object for the given task among many objects in the environment.

\begin{figure}[tb!]
\centering
\includegraphics[width=0.98\columnwidth]{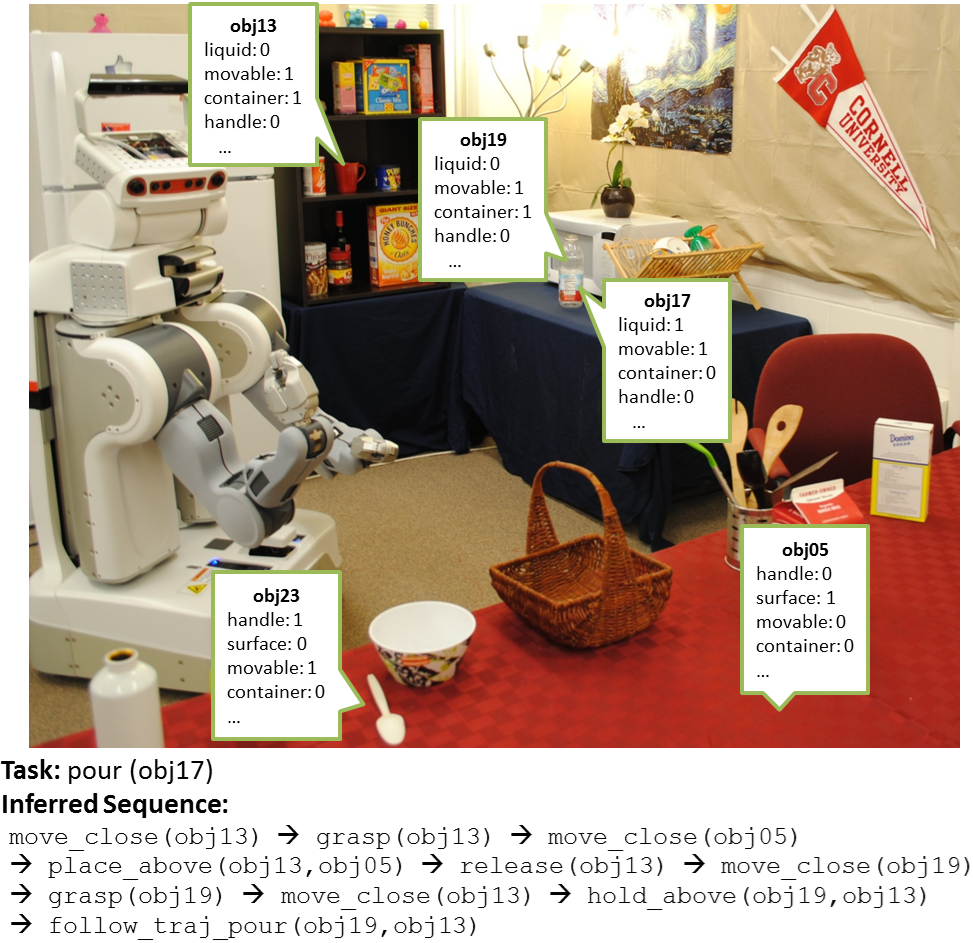}
\vspace*{\captionReduceTop}
\caption{Figure showing our Kodiak PR2 in a kitchen with different objects labeled
with attributes. To accomplish the under-defined task of \texttt{pour(obj17)},
it has to first find the mug (obj13) and carry it to the table (obj05)
since it is dangerous to pour liquid in a tight shelf. Once the mug is
on the table, it has to bring the liquid by the container (obj19) and then
finally pour it into the mug.}
\label{fig:exampletask}
\end{figure}

We take a dynamic planning approach to the problem of synthesizing, in
the right order, the suitable primitive controllers. 
The best primitive
to execute at each discrete time step is based on a score function
that represents the appropriateness of a particular primitive for the
current state of the environment.  Conceptually, a dynamic plan
consists of a loop containing a sequence of conditional statements
each with an associated primitive controller or action. If the current
environment matches the conditions of one of the conditional
statements, the corresponding primitive controller is executed,
bringing the robot one step closer to completing the overall task 
(example in Section~\ref{sec:approach}). 
We will show how to generalize sequencing of primitives to make
them more flexible and robust, by switching to an attribute-based
representation. 
We then show how to unroll the loop into a graph-based
representation, isomorphic to a Markov Random Field. 
We then train the parameters of the model by maximum margin learning
method using a dataset comprising many examples of sequences.


We evaluated our model on 127 controller sequences for
five under-specified manipulation tasks generated from 13 environments using 7
primitives.  We show that our model can predict suitable primitives to
be executed with the correct arguments in most settings. Furthermore,
we show that, for five high-level tasks, 
our algorithm was able to correctly sequence 70\% of the sequences in different environments.

The main contributions of this paper are:
\begin{packed_item}
\item using an attribute-based representation of the environment for task planning,
\item inferring the sequence of steps where the goals are under-specified and have to be inferred from the context, 
\item a graph-based representation of a dynamic plan by unrolling the loop into a Markov Random Field.
\end{packed_item}



\section{Related Work}

There is a large body of work in task planning across various communities. We describe
some of them in the following categories.

\noindent
\textbf{Manual Controller Sequencing.}
Many works manually sequence different types of controllers
to accomplish specific types of tasks.
\citet{bollini2011bakebot} develop an end-to-end system which can find ingredients
on a tabletop and mix them uniformly to bake cookies.
Others used pre-programmed sequences for tea serving and carrying humans
in healthcare robotics \cite{nakai2006development,mukai2010development}.
These approaches however cannot scale to large number of tasks when each task
requires its own complicated rules for sequencing controllers
and assumes a controlled environment, 
which is very different from actual human households, where objects of interest 
can appear anywhere in the environment
with a variety of similar objects.

\citet{beetz2011robotic} retrieve a sequence for ``making a pancake'' from online websites
but assumes an environment with correct labels and a single choice of object for the task. 
Human experts can generate finite state machines for robots but 
this again requires explicit labels (e.g. AR tags) \cite{nguyen2013ros}.
Our work addresses these problems by representing each object in the environment as a set 
of attributes which is more robust than labeling the individual object
\cite{ferrarilearning, farhadi2009describing, lampert2009learning}.
In our recent work \cite{misra_tellmedave_2014},
we learn a sequence given a natural language instruction and object labels,
where the focus is to learn the grounding of 
the natural language into the environment.

\noindent
\textbf{Learning Activities from Videos.}  
In the area of computer vision,
several works \cite{yang2010recognizing,yao2010modeling,sung_rgbdactivity_2012,koppula2013_anticipatingactivities} 
consider modeling the sequence of activities
that humans perform.  These works are complementary to ours because our problem is to infer
the sequence of controllers and not to label the videos.

\noindent
\textbf{Symbolic Planning.}
Planning problems often rely on symbolic representation of entities as well as their relations.
This has often been formalized as a deduction \cite{green1969application}
or satisfiability problem \cite{kautz1992planning}.
A plan can also be generated hierarchically by first planning
abstractly, and then generating a detailed plan recursively \cite{kaelbling2011hierarchical}. 
Such approaches can generate a sequence of controllers that can be proven
to be correct \cite{johnson2011probabilistic,belta2007symbolic}.
Symbolic planners however require encoding every precondition and effect of each operation,
which will not scale in human environments where there are large variations.
Such planners also require domain description for each planning domain 
including the types of each object (e.g., pallet crate - surface, hoist surface - locatable)
as well as any relations 
(e.g., on x:crate y:surface, available x:hoist).
The preconditions and effects 
can be learned directly from examples of recorded plans \cite{yang2007learning, zhuo2010learning}
but this method suffers when there is noise in the 
data \cite{zhuo2010learning}, and also suffers from the difficulty of
modeling real world situations with the PDDL representation \cite{yang2007learning}.

Such STRIPS-style representation also restricts the environment to be represented
with explicit labels.
Though there is a substantial body of work on labeling human environments
\cite{koppula2011semantic, lai2012detection}, it still remains a challenging task.
A more reliable way of representing an environment is representing
through attributes \cite{ferrarilearning, farhadi2009describing}.
An attribute-based representation 
even allows classification of object classes that are not present in the training data \cite{lampert2009learning}. 
Similarly, in our work, we represent the environment as a set of attributes,
allowing the robot to search for objects with the most suitable attributes
rather than looking for a specific object label.

\noindent
\textbf{Predicting Sequences.}
Predicting sequences has mostly been studied in a Markov Decision Process framework,
which finds an optimal policy given the reward for each state.
Because the reward function cannot be easily specified in many applications,
inverse reinforcement learning (IRL)
learns the reward function from an expert's policy \cite{ng2000algorithms}. 
IRL is extended to Apprenticeship Learning
based on the assumption that the expert tries to optimize an unknown reward function \cite{abbeel2004apprenticeship}.
Most similar to our work, the Max-Margin Planning
frames imitation learning as a structured max-margin learning problem \cite{ratliff2006maximum}.
However, this has only been applied to problems such as 2D path planning,
grasp prediction and footstep prediction \cite{ratliff2009learning}, which
have much smaller and clearer sets of states and actions compared to
our problem of sequencing different controllers.
Co-Active Learning for manipulation path planning \cite{jainsaxena2013_trajectorypreferences},
where user preferences are learned from weak incremental feedback,
does not directly apply to sequencing different controllers.

Both the model-based and model-free methods evaluate state-action pairs.
When it is not possible to have knowledge about all possible or subsequent states (\emph{full backup}),
they can rely on \emph{sample backup} 
which still requires sufficient sample to be drawn from the state space \cite{qualitative_rl_book}.
However, when lots of robot-object interactions are involved,
highly accurate and reliable physics-based robotic simulation is required along 
with reliable implementation of each manipulation controllers. Note that each of 
the manipulation primitives such as grasping are still not fully solved problems.
For example, consider the scenario where the robot is grasping the edge of the table and
was given the instruction of \texttt{follow\_traj\_pour(table,shelf)}.
It is unclear what should occur in the environment and becomes challenging to have reliable simulation
of actions.
Thus, in the context of reinforcement learning, we take a maximum margin
based approach to learning the weight for $\textbf{w}^T \phi(s,a)$ such that it maximizes the number of states
where the expert outperforms other policies, 
and chooses the action that maximizes $\textbf{w}^T \phi(s,a)$ at each time step.
The key in our work is representing task planning as a graph-based model and designing 
a score function that uses attribute-based representation of environment for under-specified tasks.


\section{Our Approach}
\label{sec:approach}

We refer to a sequence of \emph{primitives} (low-level navigation and manipulation controllers) 
as a \emph{program}.
To model the sequencing of primitives, 
we first represent each object in the environment with a set of attributes 
as described in Section~\ref{sec:attributes}.
In order to make programs generalizable, primitives should have the following
two properties.
First, each primitive should specialize in an atomic operation 
such as moving close, pulling, grasping, and releasing. 
Second, a primitive should not be specific to a single high-level task.
By limiting the role of each primitive and keeping it general, 
many different manipulation tasks can be accomplished with the same small set of primitives,
and our approach becomes easily adaptable to different robots by providing implementation of
primitives on the new robot. 

For illustration, we write a program for ``throw garbage away'' in Program \ref{alg:program_example}.
Most tasks could be written in such a format, where 
there are many \texttt{if} statements inside the loop. 
However, even for a simple ``throw garbage away''
task, the program is quite complex.
Writing down all the rules that can account for the many different scenarios that can arise in
a human environment would be quite challenging.

\vskip -.1in
\begin{algorithm}
	\caption{``throw garbage away.''}
	\label{alg:program_example}	
\small
\begin{algorithmic}
   	\STATE {\bfseries Input:} environment $e$, trash $a_1$
   	\STATE $gc = find\_garbage\_can(e)$
   	\REPEAT
  	\IF{$a_1$ is in hand \& $gc$ is close}
	\STATE release($a_1$)  	
  	\ELSIF{$a_1$ is in hand \& far from $gc$}
	\STATE move\_close($gc$)  	
   	\ELSIF{$a_1$ is close \& $a_1$ not in hand \\
   		  \qquad \quad \& nothing on top of $a_1$}
   	\STATE grasp($a_1$) \\
	\quad \vdots
   	\ELSIF{$a_1$ is far}
	\STATE move\_close($a_1$)
	\ENDIF
   	\UNTIL{$a_1$ inside $gc$}
\end{algorithmic}
\end{algorithm}
\vskip -.1in

Program \ref{alg:program_example} is an example of what is commonly referred to as 
reactive or dynamic 
planning \cite{ai_book,koenig}. In traditional deliberative planning, a planning algorithm synthesizes a 
sequence of steps that starts from the given state and reaches the given goal state. 
Although current symbolic planners can find optimal plan sequences consisting of hundreds of steps,
such long sequences often break down because of unexpected events during the execution. 
A dynamic plan provides a much more robust alternative. 
At each step, the current state of the environment is considered 
and the next appropriate action is selected by one of the conditional statements 
in the main loop. 
A well-constructed dynamic plan will identify the next step required to bring the robot closer
to the overall goal in any possible world state.
In complex domains, dynamic plans may become too complicated. However, 
we are considering basic human activities, such as following a recipe,
where dynamic plans are generally quite compact and can effectively lead the robot to the 
goal state. Moreover, as we will demonstrate, we can learn the dynamic plan 
from observing a series of action sequences in related environments.

In order to make our approach more general, we introduce a feature based 
representation for the conditions of \texttt{if} statements.
We can extract some features from both the environment and
the action that will be executed in the body of \texttt{if} statement.
With extracted features $\phi$ and some weight vector $\textbf{w}$ for each \texttt{if}
statement, the same conditional statements can be written as $\textbf{w}^T\phi$,
since the environment will always contain the rationale for executing certain primitive.
Such a feature-based approach allows us to re-write Program \ref{alg:program_example}
in the form of Program \ref{alg:program_example2}.

\vskip -.1in
\begin{algorithm}
	\caption{``throw garbage away.''}
	\label{alg:program_example2}
\small
\begin{algorithmic}
   	\STATE {\bfseries Input:} environment $e$, trash $a_1$
   	\STATE $gc = find\_garbage\_can(e)$
   	\REPEAT
	\STATE $e_t = $ current environment
  	\IF{$w_1^T \phi$($e_t$,release($a_1$)) $>$ 0}
	\STATE release($a_1$)  	
   	\ELSIF{$w_2^T \phi$($e_t$,move\_close($gc$)) $>$ 0}	
	\STATE move\_close($gc$) \\
	\quad \vdots
   	\ELSIF{$w_n^T \phi$($e_t$,move\_close($a_1$)) $>$ 0}	
	\STATE move\_close($a_1$)	
	\ENDIF
   	\UNTIL{$a_1$ inside $gc$}
\end{algorithmic}
\end{algorithm}
\vskip -.1in

Now all the \texttt{if} statements have the same form, where the same primitive along with same arguments
are used in both the condition as well as the body of the \texttt{if} statement. 
We can therefore reduce all \texttt{if} statements inside the loop 
further down to a simple line which depends only on a single weight vector and 
a single joint feature map, as shown in Program~\ref{alg:program_example3},
for finding the most suitable pair of primitive $\hat{p}_t$ and its arguments ($\hat{a}_{1,t}$, $\hat{a}_{2,t}$).
\vskip -.1in
\begin{algorithm}
	\caption{``throw garbage away.''}
	\label{alg:program_example3}
\small	
\begin{algorithmic}
   	\STATE {\bfseries Input:} environment $e$, trash $g_{a1}$
    	\REPEAT
	\STATE $e_t = $ current environment
 	\STATE $(\hat{p}_t, \hat{a}_{1,t}, \hat{a}_{2,t}) := 
 		\displaystyle\argmax_{p_t  \in \mathcal{P}, a_{1,t}, a_{2,t} \in \mathcal{E}}
 		 {w^T\phi(e_t, p_t(a_{1,t}, a_{2,t}))}$
 	\STATE execute $\hat{p}_t( \hat{a}_{1,t}, \hat{a}_{2,t})$
	\UNTIL{$\hat{p}_t = done$}	
\end{algorithmic}
\end{algorithm}
\vskip -.1in

The approach taken in Program \ref{alg:program_example3} also allowed removing
the function $find\_garbage\_can(e)$.
Both Program \ref{alg:program_example} and Program \ref{alg:program_example2}
require $find\_garbage\_can(e)$ which depends on
semantic labeling of each object in the environment.
The attributes of objects will allow the program
to infer which object is a garbage can without explicit encoding.

Program \ref{alg:program_example3} provides a generic representation of a dynamic plan. 
We will now discuss an 
approach to learning a set of weights.
To do so, we will employ a graph-like representation obtained by ``unrolling'' the 
loop representing discrete time steps by different layers. We will obtain a representation 
that is isomorphic to a Markov Random Field (MRF) and will use a maximum margin based 
approach to training the weight vector.
Our MRF encodes the relations between the environment, primitive and its arguments.
Our empirical results show that such  
a framework is effectively trainable with a relatively small set of example sequences. Our 
feature-based dynamic plan formulation therefore offers an effective and general 
representation to learn and generalize from action sequences, accomplishing high-level tasks 
in a dynamic environment.

\section{Model Formulation}


We are given a set of possible primitives $\mathcal{P}$ (navigation and manipulation controllers)
to work with (see Section~\ref{sec:experiments})
and an environment $\mathcal{E}$ represented by a set of attributes.
Using these primitives, the robot has to accomplish a manipulation task $g \in \mathcal{T}$.
The manipulation task $g$ is followed by the arguments $g_{a1}, g_{a2} \in \mathcal{E}$
which give a specification of the task. For example, the program ``throw garbage away''
would have a single argument which would be the object id of the object that needs
to be thrown away.

At each time step $t$ (i.e., at each iteration of the loop in Program~\ref{alg:program_example3}),
our environment $e_t$ will dynamically change, 
and its relations with the primitive is represented with a joint set of features.
These features include information about the physical and semantic properties of the
objects as well as information about their locations in the environment.

Now our goal is to predict the best primitive $p_t \in \mathcal{P}$ to execute
at each discrete time step, along with its arguments: $p_t(a_{1,t},a_{2,t})$.
We will do so by designing a score function $S(\cdot)$ that represents 
the correctness of executing a primitive in the current environment for a task.
\begin{align*}
S(g(g_{a1}, g_{a2}),e_t, &p_t(a_{1,t},a_{2,t})) = \\
&w^T \phi(g(g_{a1}, g_{a2}),e_t, p_t(a_{1,t}, a_{2,t}))
\end{align*}

\begin{figure}[tb!]
\centering
\includegraphics[width=0.37\textwidth]{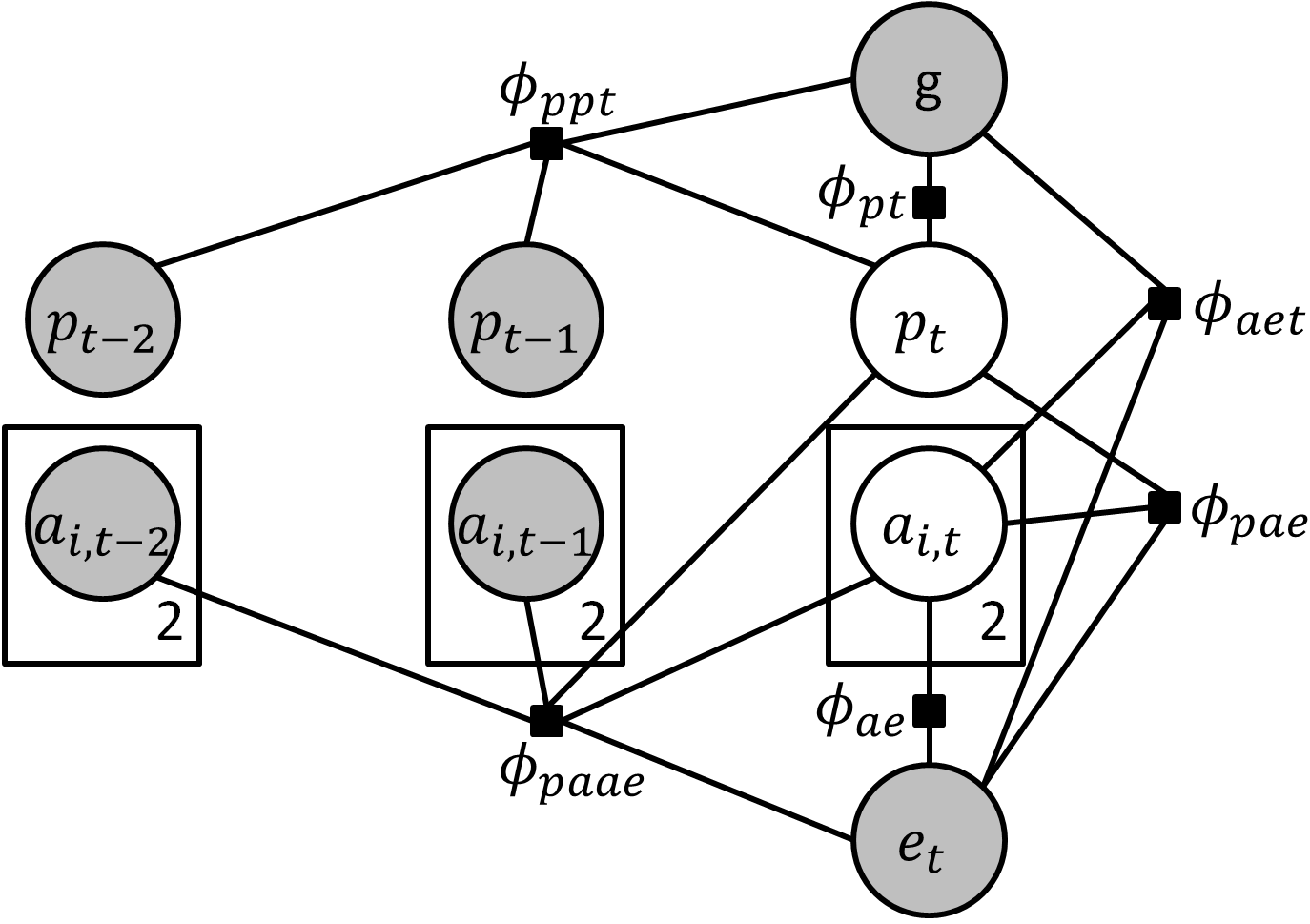}
\vspace*{\captionReduceTop}
\caption{
\textbf{Markov Random Field representation of our model} at discrete
time step $t$.
The top node represents the given task $g, g_{a1}, g_{a2}$.
The second layer from the top represents the sequence of primitives, 
and the layer below represents the arguments associated with each primitive.
And, the bottom node represents the environment represented with set of attributes.
}
\vspace*{\captionReduceBot}
\label{fig:mrf}
\end{figure}

In order to have a parsimonious representation,
we decompose our score function using a model isomorphic 
to a Markov Random Field (MRF), shown in Figure~\ref{fig:mrf}.  
This allows us to 
capture the dependency between primitives, their arguments, and environments 
which are represented by set of attributes.
In the figure, 
the top node represents the given task and its arguments ($g, g_{a1}, g_{a2}$).
The second layer from the top represents the sequence of primitives, 
and the layer below represents the arguments associated with each primitive.
And, the bottom node represents the environment which is represented with set of attributes.
Note that we also take into account the previous two primitives in the past,
together with their arguments: $p_{t-1}(a_{1,t-1}, a_{2,t-1})$ and 
$p_{t-2}(a_{1,t-2}, a_{2,t-2})$.



Now the decomposed score function is:
\begin{align*}
S = 
\underbrace{S_{ae}}_{\mathclap{\text{args-env}}} + 
\overbrace{S_{pt}}^{\mathclap{\text{prim-task}}} + 
\underbrace{S_{aet}}_{\mathclap{\text{args-env-task}}} +
\overbrace{S_{pae}}^{\mathclap{\text{prim-args-env}\quad}} + 
\underbrace{S_{ppt}}_{\mathclap{\quad\text{prim-prim(prev)-task}}} + 
\overbrace{S_{paae}}^{\mathclap{\text{\quad prim-args-args(prev)-env}}}
\end{align*}
The terms associated with an edge in the graph are defined as
a linear function of its respective features $\phi$ and weights $w$:

\vskip -.2in
{\small
\begin{align*}
S_{ae} &= {w_{ae1}}^T \phi_{ae}(a_{1,t}, e_t) + {w_{ae2}}^T \phi_{ae}(a_{2,t}, e_t) \\
S_{pt} &= {w_{pt}}^T \phi_{pt}(p_{t}, g)      
\end{align*}
}
Similarly, the terms associated with a clique in the graph are defined as
a linear function of respective features $\phi$ and weights $w$:

\vskip -.2in
{\small
\begin{align*}
S_{aet} &= {w_{aet1}}^T \phi_{aet}(a_{1,t}, e_{t}, g) + {w_{aet2}}^T \phi_{aet}(a_{2,t},e_{t},g) \\      
S_{pae} &= {w_{pae1}}^T \phi_{pae}(p_{t}, a_{1,t}, e_{t}) + {w_{pae2}}^T \phi_{pae}(p_{t}, a_{2,t},e_{t}) \\      
S_{ppt} &= {w_{ppt1}}^T \phi_{ppt}(p_{t-1}, p_{t}, g) + {w_{ppt2}}^T \phi_{ptt}(p_{t-2}, p_{t}, t) \\      
S_{paae} &= \sum_{i,j\in (1,2), k \in (t-2, t-1)}{{w_{paae_{ijk}}}^T \phi_{paae}(p_{t}, a_{i,k}, a_{j,t}, e_t)}
\end{align*}
}
Using these edge and clique terms, 
our score function $S$ can be simply written in the following form, 
which we have seen in Program \ref{alg:program_example3} with an extra term $g$ for the task:
$S(g(g_{a1}, g_{a2}),e_t,p_t(a_{1,t},a_{2,t})) = w^T \phi(g(g_{a1}, g_{a2}),e_t, p_t(a_{1,t}, a_{2,t}))$.

\subsection{Features}
\vspace*{\subsectionReduceBot}

In this section, we describe our features $\phi(\cdot)$ for the different terms in the previous section.


\emph{Arguments-environment ($\phi_{ae}$)}:  
The robot should be aware of its location and the current level of its interaction 
with objects (e.g., grasped), which are given as possible primitive arguments $a_{1,t}, a_{2,t}$.
Therefore, we add two binary features which
indicate whether each primitive argument is already grasped
and two features for the centroid distance from the robot to each primitive arguments.

For capturing spatial relation between 
two objects $a_{1,t}$ and $a_{2,t}$,
we add one binary feature indicating 
whether primitive arguments $a_{1,t}, a_{2,t}$ are currently in collision with each other.

\emph{Arguments-environment-task ($\phi_{aet}$)}:  
To capture relations between the objects of interest (task arguments) and
objects of possible interest (primitive arguments),
we build a binary vector of length 8.
First four represents the indicator values of whether the objects of interest 
are identical as the objects of possible interest, and the last four 
represents spatial relation of whether they overlap from top view.

It is important to realize the type of object that is below the objects of interests,
and the desired property (e.g., bowl-like object or table-like object)
may differ depending on the situation.
We create two feature vectors, each of length $l$.
If the robot is holding the object, we store its extracted attributes in the first vector.
Otherwise, we store them in the second vector.
If the primitive has two arguments, we use the first primitive argument
since it often has higher level of interaction with the robot
compared to the second argument.

Finally, to capture correlation between the high-level task and
the types of object in primitive argument,
we take a tensor product of two vectors:
an attribute vector of length $2l$ for two objects and
a binary occurrence vector of length $|\mathcal{T}|$.
The matrix of size $2l \times |\mathcal{T}|$ is flattened to a vector.

\emph{Primitive-task ($\phi_{pt}$)}:
The set of primitives that are useful may differ depending on the type of the task.
We create a $|\mathcal{T}|\times|\mathcal{P}|$ binary co-occurrence matrix 
between the task $g$ and the primitive $p_t$
that has a single non-zero entry in the
current task's ($g^{th}$) row and current primitive's (${p_t}^{th}$) column.

\emph{Primitive-arguments-environment ($\phi_{pae}$)}:
Some primitives such as \texttt{hold\_above} require one of the objects in arguments 
to be grasped or not to be grasped
to execute correctly.
We create a $|\mathcal{P}|\times 2$ matrix where the row for 
the current primitive (${p_t}^{th}$ row)
contains two binary values indicating  
whether each primitive argument is in the manipulator.

\emph{Primitive-primitive(previous)-task ($\phi_{ppt})$}:
The robot makes different transitions between primitives for different tasks.
Thus, a binary co-occurrence matrix of size $|\mathcal{T}|\times|\mathcal{P}|^2$ 
represents transition occurrence between the primitives for each task.
In this matrix, we encode two transitions for the current task $g$, 
from $t-2$ to $t$ and from $t-1$ to $t$.

\emph{Primitive-arguments-arguments(previous)-environment ($\phi_{paae}$)}:
For a certain primitive in certain situations, the arguments 
may not change between time steps.
For example, \texttt{pour(A,B)} would often be preceded by \texttt{hold\_above (A,B)}.
Thus, the matrix of size $|\mathcal{P}|\times 8$ is created, 
with the ${p_t}^{th}$ row containing 8 binary values representing
whether the two primitive arguments at time $t$ are the same as
the two arguments at $t-1$ or the two arguments at $t-2$. 



\subsection{Attributes.}
\label{sec:attributes}

Every object in the environment including tables and the floor
is represented using the following set of attributes: 
height $h$, max(width($w$),length($l$)), min($w, l$), volume($w*l*h$), 
min($w,l,h$)-over-max($w,l,h$), median($w,l,h$)-over-max($w,l,h$), 
cylinder-shape, box-shape, liquid, container, handle, movable, 
large-horizontal-surface, and multiple-large-horizontal-surface.
Attributes such as cylinder-shape, box-shape, container, handle, and
large-horizontal-surface can be reliably extracted from RGB or RGBD images, and were shown
to be useful in several different applications
\cite{ferrarilearning, farhadi2009describing, lampert2009learning, koppula2011semantic}.
We study the effects of attribute detection errors on our model in Section~\ref{sec:experiments}.

\begin{figure}[tb]

\centering
\includegraphics[width=0.48\columnwidth]{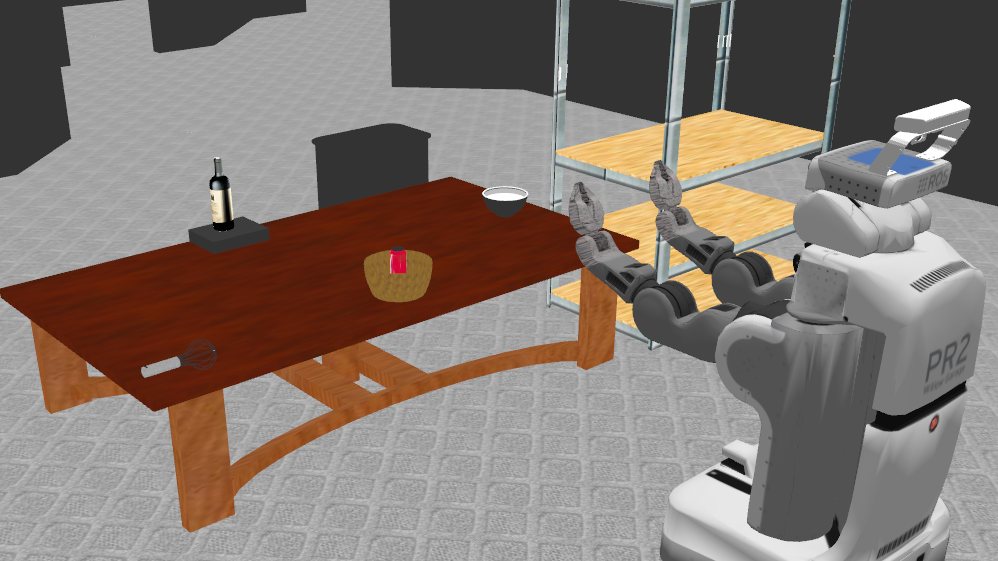}
\includegraphics[width=0.48\columnwidth]{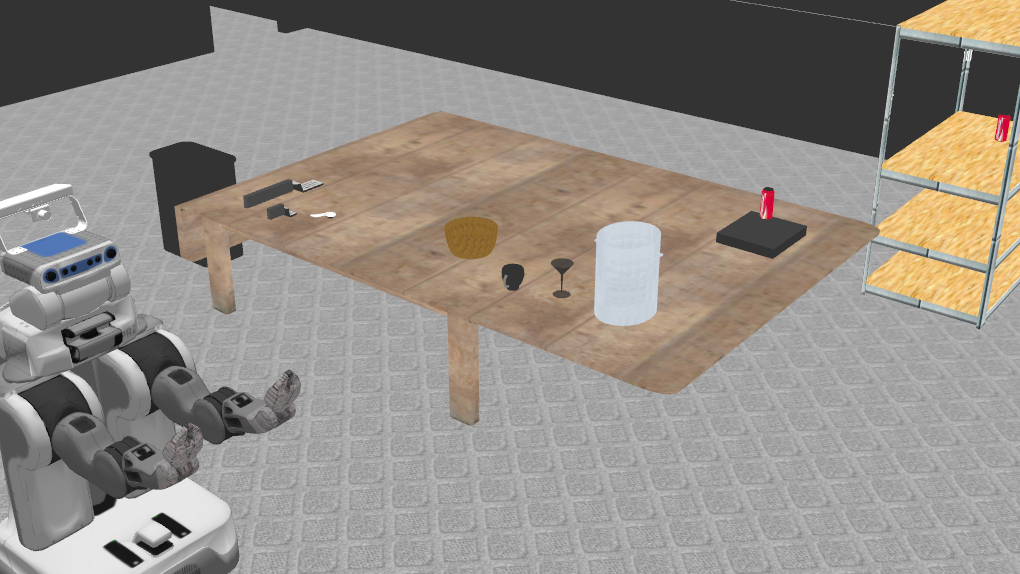}
\vspace*{\captionReduceTop}
\caption{Figure showing two of our 13 environments in our evaluation dataset
		using 43 objects along with PR2 robot.}
\label{fig:env}
\vspace*{\captionReduceBot}
\end{figure}

\subsection{Learning}
We use a max-margin approach to train a single model for all tasks.
This maximum margin approach fits our formulation, since it assumes that
the discriminant function is a linear function of a weight vector $\textbf{w}$ and
a joint feature map $\phi(g(g_{a1}, g_{a2}),e_t, p_t(a_{1,t}, a_{2,t}))$, and it 
has time complexity linear with the number of training examples when
solved using the cutting plane method \cite{joachims2009cutting}.
We formalize our problem as a ``1-slack'' structural SVM optimization problem:

{\small
\vskip -.2in
\begin{align*}
&\qquad \min_{\textbf{w}, \xi  \geq 0}{\frac{1}{2} \textbf{w}^T\textbf{w} + \frac{C}{l} \sum_{i=1}^{n}\sum_{t=1}^{l^i}\xi^i_t}\\
&s.t. \quad \text{for} \;\; 1\leq i \leq n, \text{for each time step} \; t: \;\;\;
\forall \hat{p} \in \mathcal{P},   \forall \hat{a}_1,\hat{a}_2 \in \mathcal{E}: \\
&\mathbi{w}^T {[\phi(g^i(g^i_{a1}, g^i_{a2}),e^i_t, p^i_t(a^i_{1,t}, a^i_{2,t}))\!-\!\phi(g^i(g^i_{a1}, g^i_{a2}),e^i_t, \hat{p} (\hat{a}_{1}, \hat{a}_{2}))]} \\
&\quad  \geq \Delta(\{ p^i_t, a^i_{1,t}, a^i_{2,t}\}, \{\hat{p}, \hat{a}_1, \hat{a}_2\}) - \xi^i_t
\end{align*}
\vskip -.1in
}

\noindent
where $n$ is the number of example sequences, $l^i$ is the length of the $i^{th}$ sequence,
and $l$ is the total length combining all sequences.
The loss function is defined as: 

{\small
\vskip -.2in
\begin{align*}
\Delta(\{p, a_1, a_2\}, \{\hat{p}, \hat{a}_1, \hat{a}_2\})
& = \mathds{1}(p \neq \hat{p}) \!+\! \mathds{1}(a_1 \neq \hat{a}_1)
 \!+\! \mathds{1}(a_2 \neq \hat{a}_2) 
\end{align*}
\vskip -.05in
}

With a learned $\textbf{w}$, we choose the next action in sequence
by selecting a pair of primitive and arguments that gives the largest discriminant value:
$$\argmax_{p_t  \in \mathcal{P}, a_{1,t}, a_{2,t} \in \mathcal{E}}
 		 {w^T \phi(g(g_{a1}, g_{a2}),e_t, p_t(a_{1,t}, a_{2,t}))}$$


\begin{table*}
\begin{center} 
\caption{
Result of baselines, our model with variations of feature sets, and our full model
on our dataset consisting of 127 sequences.
The ``prim'' columns represent percentage of primitives correctly chosen
regardless of arguments, and ``args'' columns represent percentage of a correct pair of
primitive and arguments. 
The last column shows average percentage of sequences correct over the five
programs evaluated.
}
\resizebox{\linewidth}{!}{
\begin{tabular}
{@{}l|cc|cc|cc|cc|cc|cc|cc|cc|cc@{}}
\hline 
& \multicolumn{2}{c|}{move\_close} & \multicolumn{2}{c|}{grasp} 
& \multicolumn{2}{c|}{release} & \multicolumn{2}{c|}{place\_above}
& \multicolumn{2}{c|}{hold\_above}& \multicolumn{2}{c|}{traj\_circle}
& \multicolumn{2}{c|}{traj\_pour} & \multicolumn{2}{c|}{\textbf{Average}} 
& \multicolumn{2}{c}{\textbf{Sequence}} \\ 
 \hline
 & prim & arg  & prim & \;arg  & prim & arg  & prim & arg  & prim & arg 
 & prim & arg  & prim & arg  & prim & arg  & prim & arg\\
 \hline
\textit{chance} &
14.3 & 1.1 & 14.3 & 1.1 & 14.3 & 1.1 & 14.3 & 0.1 & 14.3 & 0.1 & 14.3 & 1.1 & 14.3 & 0.1 & 14.3 & 0.7 & 0 & 0 \\
\textit{multiclass} &
99.6 & - & 90.4 & - & 95.7 & - & 68.5 & - & 79.7 & - & 100.0 & - & 14.7 & - & 78.4 & - & - & - \\
\hline
\textit{symb-plan-svm} &
99.6 & 82.5 & 94.2 & 72.4 & 67.4 & 63.0 & 60.9 & 43.5 & 76.6 & 73.4 & 96.7 & 76.7 & 97.1 & 91.2 & 84.6 & 71.8 & 58.4 & 49.6 \\
\textit{symb-plan-manual} &
99.6 & 85.4 & 94.2 & 76.3 & 67.4 & 63.0 & 60.9 & 50.0 & 76.6 & 76.6 & 96.7 & 96.7 & 97.1 & 97.1 & 84.6 & 77.9 & 58.4 & 54.9 \\
\hline
\textit{Only edge features} &
23.5 & 15.3 & 56.4 & 45.5 & 93.5 & 93.5 & 0.0 & 0.0 & 18.8 & 9.4 & 100.0 & 100.0 & 50.0 & 44.1 & 48.9 & 44.0 & 0 & 0\\
\textit{Only clique features} &
99.6 & 1.9 & 96.8 & 82.7 & 90.2 & 90.2 & 72.8 & 15.2 & 87.5 & 15.6 & 96.7 & 96.7 & 100.0 & 97.1 & 91.9 & 57.0 & 45.0 & 0 \\
\textit{\textbf{Ours - full}} &
99.3 & 82.8 & 96.8 & 84.0 & 97.8 & 97.8 & 89.1 & 79.3 & 96.9 & 92.2 & 100.0 & 100.0 & 97.1 & 94.1 & 96.7 & \textbf{90.0} & 91.6 & \textbf{69.7}\\

\hline
\end{tabular} 
}
\label{tab:args}
\end{center}
\vskip -.22in
\end{table*}

\section{Experiments}
\label{sec:experiments}

\noindent
\textbf{Dataset.}
We considered seven primitives (low-level controllers):
\texttt{move\_close (A)}, \texttt{grasp (A)}, \texttt{release (A)},
\texttt{place\_above (A,B)}, \texttt{hold\_above (A,B)},
\texttt{follow\_traj\_circle (A)} and \texttt{follow\_traj\_pour (A,B)}.
Depending on the environment and the task, these primitives could
be instantiated with different arguments.  For example,
consider an environment that contains a bottle (obj04) containing liquid (obj16)
and an empty cup  (obj02) placed on top of the shelf, among other objects.
If, say from a recipe, our task is to pour the liquid, then our program
should figure out the correct sequence of primitives with correct arguments
(based on the objects' attributes, etc.):

{\scriptsize
\vskip -.2in
\begin{align*}
\{&\texttt{pour(obj16)};env2\} \rightarrow \\
\{&\texttt{move\_close(obj02); grasp(obj02); move\_close(obj04);}\\
&\texttt{place\_above(obj02,obj26); release(obj02); grasp(obj04);}\\
&\texttt{hold\_above(obj04,obj02); follow\_traj\_pour(obj04,obj02)}\}
\end{align*}
}
Note that the actual sequence does not directly interact with the liquid (obj16)---the only object specified by the task---but
rather with a container of liquid (obj04), an empty cup (obj02), and
a table (obj26), while none of these objects are specified in the task arguments. 
As seen in this example, the input for our planning problem is  under-specified.

For evaluation, we prepared a dataset where the goal was to produce correct sequences
for the following tasks in different environments:
\begin{packed_item}

\item \texttt{stir(A)}: 
Given a liquid A, the robot has to identify a stirrer of ideal size 
(from several) and stir with it. 
The liquid may be located on a tight shelf
where it would be dangerous to stir the liquid, and the robot should always stir it
on top of an open surface, like a table.
The robot should always
only interact with the container of the liquid, rather 
than the liquid itself, whenever liquid needs to be carried or poured. 
Our learning algorithm should learn such properties.

\item \texttt{pick\_and\_place(A,B)}:
The robot has to place A on top of B.
If A is under some other object C, the object C must first be moved 
before interacting with object A.

\item \texttt{pour(A)}:
The robot has to identify a bowl-like object without object labels and pour liquid A into it.
Note again that liquid A cannot be directly interacted with, 
and it should not be poured on top of a shelf.

\item \texttt{pour\_to(A,B)}:
The liquid A has to be poured into the container B. (A variant of the previous task
where the container B is specified but the model should be able to distinguish two different tasks.)

\item \texttt{throw\_away(A)}:
The robot has to locate a garbage can in the environment and throw out object A.
\end{packed_item}

In order to learn these programs, 
we collected 127 sequences for 113 unique scenarios by presenting participants the environment in simulation
and the task to be done.
We considered a single-armed
mobile manipulator robot for these  tasks.
In order to extract information about the environment at each time frame of every sequence,
we implemented each primitive using OpenRAVE simulator \cite{diankov_thesis}.
Though most of the scenarios 
had a single optimal sequence, multiple sequences were introduced when there
were other acceptable variations. The length of each sequence varies from 4 steps to 10 steps, 
providing a total of 736 instances of primitives.
To ensure variety in sequences, sequences were generated based on the 13 different environments 
shown in Figure~\ref{fig:env}, using 43 objects each with unique attributes.

\begin{figure*}[t]
\subfigure[\textbf{Confusion matrix} for the seven primitives in our dataset.
        Our dataset consist of 736 instances of seven primitives in 127 sequences 
        on five manipulation tasks.]{
\includegraphics[width=0.29\textwidth, height=1.45in]{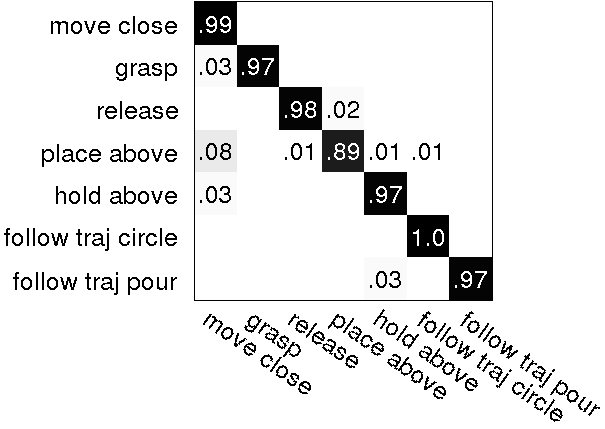}
\label{fig:confmat}
}
\;
\subfigure[\textbf{Percentage of programs correct.} Without any feedback in completely
autonomous mode, the accuracy is 69.7\%.  With feedback (number of 
feedbacks on x-axis), the performance increases.  This is on full
127 sequence dataset.]{
\includegraphics[width=0.30\textwidth, height=1.45in]{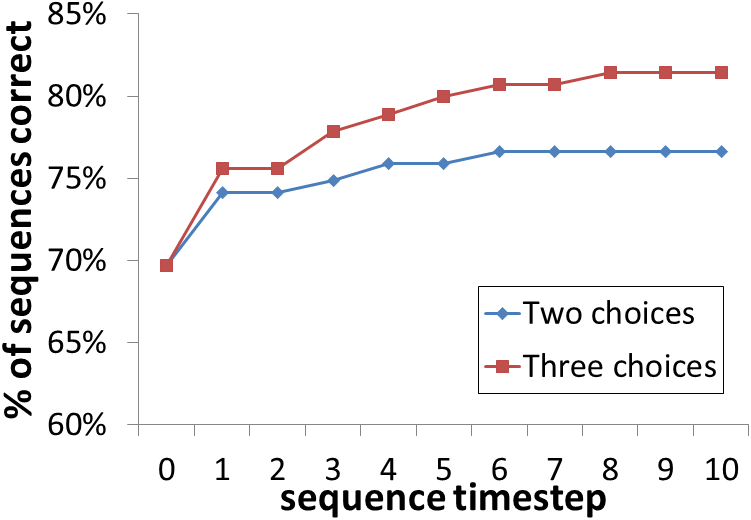}
\label{fig:feedback1}
}
\;
\subfigure[\textbf{Percentage of programs correct for 12 high-level tasks} such as making sweet tea. 
In completely
autonomous mode, the accuracy is 75\%.  With feedback (number of 
feedbacks on x-axis), the performance increases.]
{\includegraphics[width=0.30\textwidth, height=1.45in]{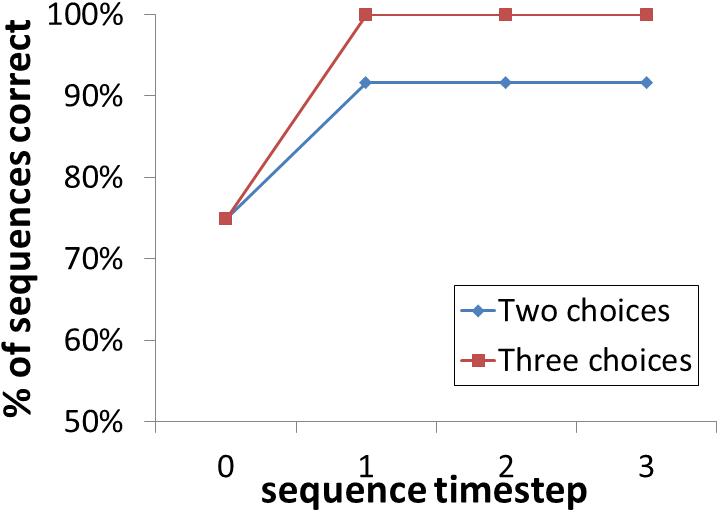}
\label{fig:feedback2}
}
\vspace*{\captionReduceTop}
\caption{\textbf{Results with cross-validation.} 
(a) On predicting the correct primitive individually.
(b) On predicting programs, with and without user intervention. 
(c) On performing different tasks with the predicted sequences.
}
\vspace*{\captionReduceBot}
\label{fig:graphs}
\end{figure*}

\noindent
\textbf{Baseline Algorithms.}
We compared our model against following baseline algorithms:
\begin{packed_item}
\item \textit{chance}: 
At each time step, a primitive and its arguments are selected at random.

\item \textit{multiclass}:
A multiclass SVM \cite{joachims2009cutting} was trained to predict primitives 
without arguments, since the set of possible arguments changes depending on the environment.

\item \textit{symbolic-plan-svm}: 
A PDDL-based symbolic planner \cite{yang2007learning, zhuo2010learning} 
requires a domain and a problem definition.
Each scenario was translated to symbolic entities and relations. However,
the pre-conditions and effects of each action in domain definition were hand-coded,
and each object was labeled with attributes using predicates.
Unlike our model that works on an under-specified problem,
each symbolic planning problem requires an explicit goal state.
In order to define these goal states, we have trained 
ranking SVMs \cite{joachims2006training} in order to 
detect a `stirrer', an `object to pour into' and a `garbage can'
for \texttt{stir}, \texttt{pour}, and \texttt{throw\_away}, respectively.
Each symbolic planning instance was then solved by reducing to a satisfiability problem
\cite{kautz1992planning, rintanen2012planning}.
 
\item \textit{symbolic-plan-manual}: 
Based on the same method as \textit{symbolic-plan-svm}, instead of training ranking SVMs,
we provided ground-truth goal states.
Even after providing lots of hand-coded rules, it is still missing some rules
due to the difficulty of representation using PDDL \cite{yang2007learning, zhuo2010learning}, 
These missing rules include the fact that liquid needs to be handled 
through its container and 
that objects should not be manipulated on top of the shelf.

\end{packed_item}

\noindent
\textbf{Evaluation and Results.}
We evaluated our algorithm through \emph{6-fold cross-validation}, computing accuracies
over primitives, over primitives with arguments, and over the full sequences.
Figure \ref{fig:confmat} shows the confusion matrix for prediction 
of our seven primitives. We see that our model is quite robust for most primitives.

With our dataset, our model was able to correctly predict pairs of primitives and arguments
90.0\% of the time and full sequences 69.7\% of the time (Table \ref{tab:args}).
Considering only the primitives without arguments, 
it was able to predict primitive 96.7\% of the time and full sequence 91.6\% of the time.
The last column of Table \ref{tab:args} shows the performance with respect to whether the complete
sequence was correct or not.
For example, for ``pouring'', our model has learned not only to bring a cup over to
the table, but also to pick out the cup when there are multiple other objects like a pot, a bowl, or a can
that may have similar properties.

\textbf{How do baselines perform for our under-specified planning problem?}
The results of various baseline algorithms are shown in Table \ref{tab:args}.
If the primitive and arguments pairs are predicted at random, 
none of the sequences would be correct because of the large search space of arguments.
\textit{Multiclass} predicted well for some of the primitives but
suffered greatly on primitives like  
\texttt{place\_above}, \texttt{hold\_above} and \texttt{follow\_traj\_pour},
which drastically impacts constructing overall sequences, even with correct arguments
selected.

The symbolic planner based approaches, \textit{symbolic-plan-svm} and \textit{symbolic-plan-manual},
suffered greatly from under-specified nature of the problem.
The planners predicted correctly 49.6\% and 54.9\% of the times, respectively, 
compared to our model's performance of 69.7\%.
Even though both planners made use of heavily hand-coded domain definitions of the problem,
due to the nature of the language used by symbolic planners, rules such as that liquid should not be 
handled on top of shelves were not able to be encoded.
Even if the language were capable of encoding these rules, it would require a human expert 
in planning language to carefully encode every single rule the expert can come up with.

Also, by varying the set of features, 
it is evident that without very robust primitive-level accuracies, 
the models are unable to construct a single correct sequence.

\begin{figure}[tb!]
\vskip -3mm
 \centering
\includegraphics[width=0.9\columnwidth]{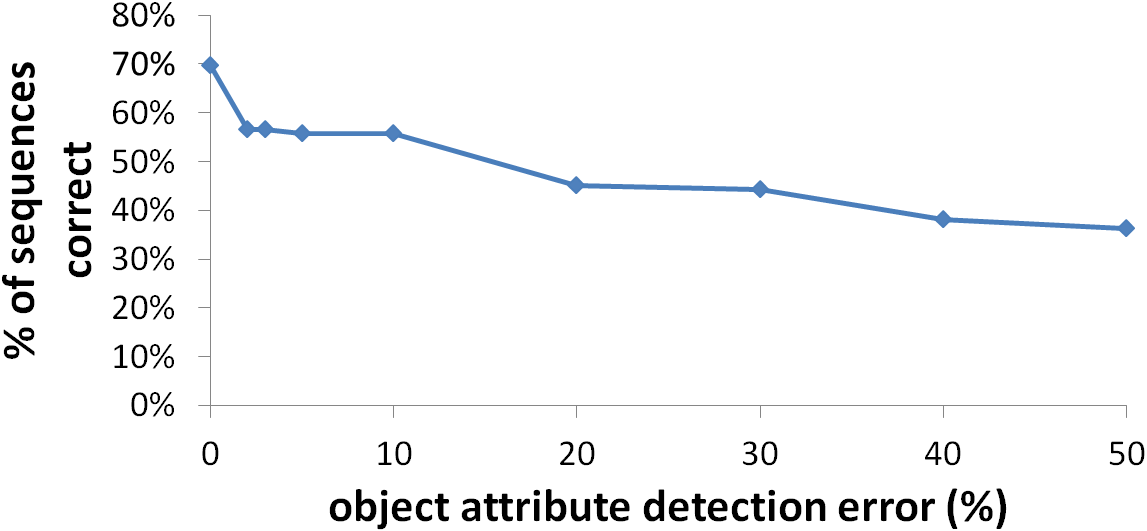}
\vspace*{\captionReduceTop}
\caption{\textbf{Effect of attribute perception error.}
Figure showing percentage of programs correct with attribute labeling errors
for binary attributes. 
For 113 unique scenarios, binary attributes were randomly flipped.}
\vspace*{\captionReduceBot}
\vspace*{\captionReduceBot}
\label{fig:artnoise}
\end{figure}

\textbf{How important is attribute representation of objects?} 
For 113 unique scenarios in our dataset,
we have randomly flipped binary attributes and observed the effects of detection errors
on correctness for the full sequence (Figure~\ref{fig:artnoise}).
When there is no error in detecting attributes, our model performs at 69.7\%.
With 10\% detection error, it performs at 55.8\%, 
and with 40\% detection errors, it performs at 38.1\%.
Since the attribute detection is more reliable than the object detection 
\cite{ferrarilearning, farhadi2009describing, lampert2009learning},
our model will perform better than planners based on explicit object labels.


\begin{figure*}[tbh!]
\vskip -.05in
\centering
\includegraphics[width=0.95\textwidth,height=34mm]{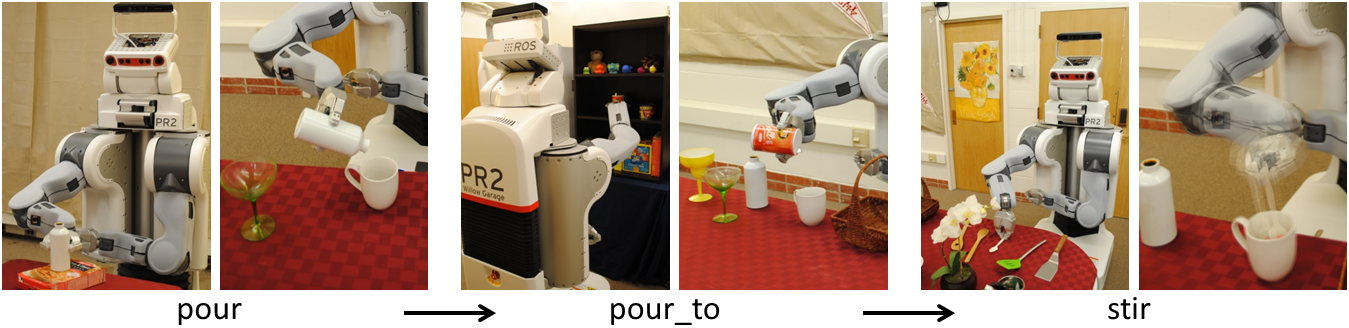}
\vspace*{\captionReduceTop}
\caption{\textbf{Few snapshots of learned sequences} forming the higher level
task of serving sweet tea, which takes the sequence of pouring tea into a cup,
pouring sugar into a cup, and then stirring it.}
\label{fig:exampletask}
\vspace*{\captionReduceBot}
\end{figure*}

\textbf{How can the robot utilize learned programs?}
These learned programs can form higher level tasks such as making
a recipe found online. For example, serving sweet tea would require the
following steps: pouring tea into a cup, pouring sugar into a cup, and stirring it
(Figure~\ref{fig:exampletask}). 
We have tested each of the four tasks, 
\emph{serve-sweet-tea}, \emph{serve-coffee-with-milk},
\emph{empty-container-and-throw-away}, and \emph{serve-and-store},
in three environments.
Each of the four tasks can be sequenced in following manner by programs respectively:
\texttt{pour} $\rightarrow$ \texttt{pour\_to} $\rightarrow$ \texttt{stir},
\texttt{pour\_to} $\rightarrow$ \texttt{pour\_to},
\texttt{pour} $\rightarrow$ \texttt{throw\_away},
and \texttt{pour} $\rightarrow$ \texttt{pick\_and\_place}.
Out of total 12 scenarios, our model was able to
successfully complete the task for 9 scenarios.

\textbf{Does the robot need a human observer?}
In an assistive robotics setting, a robot will be accompanied by a human 
observer. With help from the human, performance can be greatly improved.
Instead of choosing a primitive and argument pair
that maximizes the discriminant function, the robot can present the top 2 or 3 
primitive and argument pairs to the observer, who can simply give feedback on the
best option among those choices. 
At the initial time step of the sequence,
with only a single piece of feedback, given 2 or 3 choices,
performance improves to 74.1\% and 75.6\% respectively from 69.7\% 
(Figure \ref{fig:feedback1}).
If feedback was provided through whole sequence with the top 2 or 3 choices,
it further improves to 76.7\% and 81.4\%.
Furthermore, the four higher level tasks (recipes) considered earlier
also shows that with a single feedback at the initial time step of each program,
the results improve from 75\% to 100\% (Figure \ref{fig:feedback2}).

\smallskip
\noindent
\textbf{Robotic Experiments.}
Finally, we demonstrate that our inferred programs 
can be successfully executed on our Kodiak PR2 robot
for a given task in an environment.
Using our implementation of the primitives discussed in Section~\ref{sec:experiments},
we show our robot performing the task of ``serving sweet tea." It comprises
executing three programs in series -- \texttt{pour}, \texttt{pour\_to} and \texttt{stir} --
which in total required sequence of 20 primitives with correct arguments. 
Each of these programs (i.e., the sequence of primitives and arguments) is inferred 
for this environment. Figure~\ref{fig:exampletask} shows a few snapshots and the full video is 
available at:\\
\texttt{\small http://pr.cs.cornell.edu/learningtasksequences}

\section{Conclusion}

In this paper, we considered the problem of learning sequences of
controllers for robots in unstructured human environments.  
In an unstructured environment, even a simple task such as pouring
can take variety of different sequences of controllers 
depending on the configuration of the environment.
We took a
dynamic planning approach, where we represent the current state of the
environment using a set of attributes. 
To ensure that our dynamic plans are as general and flexible as possible,
we designed a score function that captures relations between task, 
environment, primitives, and their arguments, 
and we trained a set of parameters weighting the various attributes from example sequences. 
By unrolling the program, we can obtain a Markov Random
Field style representation, and use a maximum margin learning
strategy. We demonstrated on a series of example sequences that our
approach can effectively learn dynamic plans for various complex
high-level tasks.

\section*{Acknowledgements}
This work was supported in
part by ONR Grant N00014-14-1-0156, and Microsoft Faculty
Fellowship and NSF Career award to Saxena.

{
\small
\bibliographystyle{abbrvnat}
\bibliography{writeup}
}

\end{document}